\def\tmlrsubmission{1}
\def\tmlrpreprint{1}
\ifdefined\tmlrsubmission
  \documentclass[10pt]{article}
  \ifdefined\tmlrpreprint
    \usepackage[preprint]{tmlr}
  \else
    \usepackage{tmlr}
  \fi
  \usepackage{hyperref}
  \usepackage{url}
\else
  \documentclass[11pt]{article}
  \ifdefined\anonsubmission
    \usepackage[review]{acl}
  \else
    \usepackage[final]{acl}
  \fi
  \usepackage{fontspec}
  \IfFontExistsTF{Times New Roman}{\setmainfont{Times New Roman}}{\setmainfont{TeX Gyre Termes}}
  \IfFontExistsTF{PT Sans}{\setsansfont{PT Sans}}{\setsansfont{TeX Gyre Heros}}
  \IfFontExistsTF{Andale Mono}{\setmonofont{Andale Mono}}{\setmonofont{lmmono10-regular.otf}}
\fi
\usepackage{microtype}
\usepackage{graphicx}
\usepackage{booktabs}
\usepackage{tabularx}
\usepackage{array}
\usepackage{amsmath}
\newcommand{\dataset}{Turkish MMLU Pro}
\newcommand{\code}[1]{\texttt{#1}}
\title{\dataset: Traceable Option Augmentation and Its Validity Limits in Turkish Multiple-Choice Evaluation}
\ifdefined\tmlrsubmission
  \author{\name M. Ali Bayram \email malibayram20@gmail.com \\
    \addr Department of Computer Engineering, Yıldız Technical University\\
    İstanbul, Turkey}
  \ifdefined\tmlrpreprint
    \hypersetup{hidelinks,pdfauthor={M. Ali Bayram},pdftitle={Turkish MMLU Pro: Traceable Option Augmentation and Its Validity Limits in Turkish Multiple-Choice Evaluation}}
  \else
    \hypersetup{hidelinks,pdfauthor={},pdftitle={Turkish MMLU Pro: Traceable Option Augmentation and Its Validity Limits in Turkish Multiple-Choice Evaluation}}
  \fi
\else\ifdefined\anonsubmission
  \author{Anonymous submission}
  \hypersetup{pdfauthor={},pdftitle={Turkish MMLU Pro: Traceable Option Augmentation and Its Validity Limits in Turkish Multiple-Choice Evaluation}}
\else
  \author{M. Ali Bayram\\
    Department of Computer Engineering, Yıldız Technical University\\
    İstanbul, Turkey\\
    \texttt{malibayram20@gmail.com}\\
    \href{https://orcid.org/0000-0003-1298-4521}{ORCID: 0000-0003-1298-4521}}
  \hypersetup{pdfauthor={M. Ali Bayram},pdftitle={Turkish MMLU Pro: Traceable Option Augmentation and Its Validity Limits in Turkish Multiple-Choice Evaluation}}
\fi\fi

\begin{document}
\maketitle
\ifdefined\tmlrsubmission
\footnotetext{Generative AI was used for option-identifier selection during dataset construction and for assistance with manuscript review, literature checks, diagnostic analysis code, and audit preparation. The author verified the text, analyses, and references and accepts responsibility for the work. Section~\ref{sec:ethics} gives the full disclosure, including the limits of the reviewer-tool records.}
\fi

\begin{abstract}
Adding answer options can lower multiple-choice scores without improving assessment validity. Turkish MMLU Pro examines this distinction using 12,000 Turkish-source questions across 58 sections. Each question retains its stem, five original options and source key, and receives five options copied from other questions in the same section. Sentence-embedding retrieval proposes candidates; a language model selects existing identifiers. Deterministic verification reconstructs all 60,000 additions. A 25-model calibration exposes scoring and generation-budget effects. Five evaluations produce source-key accuracies of 34.8\%--81.4\%. On 981 shared questions, one API-served model falls from 93.7\% with five choices to 83.1\% with ten; 102 of 115 lost correct responses select borrowed options. The decrease is 24.4 percentage points on heuristically flagged negative stems and 5.9 points elsewhere. A completed human-checked audit of 200 sampled questions, with undocumented reviewer tool use, yields 47 and 31 multiple-answer judgments across the two record sets, 25 of the latter unresolved. These records support concern about ambiguity, while their dependence and incomplete reviewer-method documentation limit validation. Because order and labels also change, the paired comparison measures augmentation as implemented. The contribution is a traceable construction and an analysis of its validity limits, not evidence that lower ten-choice scores measure knowledge better.
\end{abstract}

\section{Introduction}

Multiple-choice evaluation depends on the answer options and scoring procedure as well as the question. Option order can change model predictions, and generated answers can disagree with first-token probability scores \citep{order,firsttoken}. Increasing the number of options lowers the uniform-guessing reference rate, but additional options are useful only if they preserve a defensible key and discriminate relevant knowledge. Educational research cautions against equating more options with better assessment \citep{rodriguez,gierl}.

Turkish evaluation includes curriculum-based TurkishMMLU \citep{turkishmmlu}, the broader five-choice TR-MMLU \citep{trmmlu_siu}, translated MMLU-Pro-TR \citep{mmluprotr}, and multi-task suites such as TurkBench and CETVEL \citep{turkbench,cetvel}. This work investigates a different construction choice: expanding an existing Turkish question collection by borrowing option text from the same source. This preserves inspectable provenance and avoids generating new wording, but an option that is incorrect for its donor need not be incorrect for its recipient.

This work analyses option augmentation and its validity limits. The central finding is a limitation of the construction itself: source-key agreement falls most on stems flagged as negative or exception questions, where borrowed options can introduce additional defensible answers. The evidence combines a 25-model calibration study, five complete runs, a paired comparison on 981 shared questions and a completed 200-question audit with explicit reconciliation and missingness, separating deterministic provenance checks, source-key scores and fallible human-checked semantic judgments. The accompanying resource, \dataset{}, comprises 12,000 questions with ten options across 58 near-balanced sections and an auditable protocol. It is offered as an instrument for studying this failure mode, not as a validated ten-choice benchmark: these scores are not proposed for ranking models.

\section{Related Work}\label{sec:related}

TurkishMMLU contains 9,807 test questions and 225 development questions across nine high-school subjects, with student correctness information that supports difficulty analysis \citep{turkishmmlu}. TR-MMLU evaluates 6,200 questions across 62 sections \citep{trmmlu_siu,trmmlu}. The source collection is the larger collection behind that lineage; the present benchmark overlaps with TR-MMLU and is not an independent test set relative to it. TUMLU also draws its Turkish questions from TurkishMMLU and converts questions to four options \citep{tumlu}. Thus, existing Turkish resources are neither uniformly five-choice nor necessarily disjoint.

MMLU-Pro-TR translates English MMLU-Pro into Turkish \citep{mmluprotr}. TurkBench and CETVEL cover broader task families, including capabilities beyond standalone knowledge questions \citep{turkbench,cetvel}. These resources complement the narrower setting examined here. The term \emph{Turkish-source} denotes stems taken from a Turkish collection rather than translated from MMLU-Pro for this study; available metadata does not establish each upstream item's original authorship language.

MMLU-Pro combines question filtering, expanded options, and expert review \citep{mmlupro}. The approach used in this work does not reproduce that pipeline: it neither filters questions by model difficulty nor generates new option text. Mining item banks for distractors has prior precedent \citep{baldwin}, while surveys distinguish candidate generation from expert and response-based evaluation \citep{distractor_survey,evaluation_survey}. Same-section retrieval and model selection are engineering choices here, not evidence that the resulting options are uniquely incorrect.

MMLU-Redux documents unclear questions, wrong keys, and missing or multiple valid answers \citep{redux}. These categories are directly relevant to augmentation. A source key can remain unchanged while the addition of another valid answer makes single-key scoring misleading. The strict scorer used in this work addresses a separate issue: reproducibly extracting an explicit answer from generated text. Accurate extraction cannot compensate for an invalid item.

The Distractor Assessment Framework separates incorrectness, plausibility, and diversity using automated reading-comprehension models \citep{daf}. Such measures offer candidate-screening baselines, but their transfer to Turkish subject questions would require validation against qualified reviewers. Large-option evaluation on Korean orthography uses repeated resampling, shuffling, and padding controls to investigate presentation effects \citep{large_options}. The present study instead has one saved presentation per condition. It adopts descriptive position diagnostics, while reserving causal attribution for controlled reevaluation.

\section{Dataset Construction}\label{sec:construction}

\subsection{Source and structural filtering}

\ifdefined\anonsubmission
The source is a public Turkish question collection at revision \code{30c94d45...cc91}; its repository identifier and direct citation are withheld for double-blind review. Its records contain section, topic, question, a zero-based answer index, explanation, and five choices.
\else
The source is \code{alibayram/turkish\_mmlu} at revision \code{30c94d45...cc91} \citep{source}. Its records contain section, topic, question, a zero-based answer index, explanation, and five choices.
\fi
Stable question identifiers are the first 24 hexadecimal characters of SHA-256 over the source revision, split, and row position. Explanations are not included in evaluation prompts.

The pooled source has 279,302 records from its train, test, and mmlu splits. Structural filtering removes malformed records, repeated normalized choices, option-dependent formulations such as ``all of the above'', likely missing visual references, duplicate normalized stems, and duplicate groups with conflicting answer texts. One deterministic owner is retained for repeated stems. This leaves 232,963 usable records (Table~\ref{tab:source}); \emph{usable} denotes these automated checks, not expert approval. The original split names describe provenance, not independent training and testing partitions for the construction described here.

\begin{table}[t]
\centering

\begin{tabular}{@{}lr@{}}\toprule
Item & Count\\\midrule
\makeatletter Source split: train & 232,136 \\
Source split: test & 40,966 \\
Source split: mmlu & 6,200 \\
Raw total & 279,302 \\
Automatically excluded & 46,339 \\
Usable source records & 232,963 \\
Selected benchmark records & 12,000 \\
 \bottomrule\makeatother
\end{tabular}
\caption{Recomputed source audit and selected-set size.}
\label{tab:source}
\end{table}

Sections with fewer than 200 usable questions are excluded. Seed 42 controls question queues and quota tie-breaking; the target is allocated as evenly as capacity permits. Sections unable to supply a quota are dropped and capacity is redistributed. The selected set contains 58 sections with 206 or 207 questions each. Selection is therefore conditional on structural eligibility and successful construction, rather than a representative sample of Turkish examinations or language use.

\subsection{Retrieval and forced selection}\label{sec:selection}

Pipeline version 6 embeds question stems using \texttt{paraphrase-\allowbreak multilingual-\allowbreak mpnet-\allowbreak base-v2} at revision \code{4328cf...11f5}. Sentence embeddings support candidate retrieval \citep{sbert}; they do not verify answer correctness. Up to 60 same-section donor questions interleave exact topic-label matches with nearest embedding neighbors. Options are collected round-robin from donors, excluding the recipient itself, canonical duplicates, option-dependent text, and simple answer-format mismatches. The candidate pool is capped at 100 options and 16,000 option characters.

One Ollama call to \code{gemma4:26b-mlx} selects exactly five distinct candidate identifiers. It uses seed 42, temperature 0, a 256-token output ceiling, a 16,384-token context setting, and \code{think=false}. The instruction retains the source key, prioritizes avoiding alternative correct answers and semantic duplicates, and then prefers answer-type and topic fit. It forbids creating or editing text. Crucially, it requires five selections even when candidates are weak. Malformed selections skip a question, and three consecutive malformed replies stop the run after checkpointing; candidate shortages and transport failures are explicit failures.

The forced-selection policy supplies a fixed option count but lacks a semantic rejection mechanism. A valid identifier response establishes only that five candidates passed structural checks. It does not establish their plausibility or incorrectness. The selector weights also appear in the evaluation roster, so construction-model dependence is disclosed rather than assumed absent.

\subsection{Export and verification}

Each accepted question combines its five original and five borrowed options. A deterministic permutation derived from its identifier and seed shuffles the ten options, and the source key is remapped to its displayed position. Verification reconstructs each record and checks exact source text, donor split/row/option location, donor section, answer remapping, and duplicate constraints. Display prompts strip surrounding whitespace; stored text is retained.

All 12,000 selected records and 60,000 additions pass reconstruction. The recorded 12,022 attempted questions and 12,000 acceptances describe pipeline throughput, not a quality rate. Borrowed options average 18.13 characters compared with 24.08 for original options (medians 13 and 16). Length may provide an origin cue. The displayed gold counts range from 1,131 to 1,273 across A--J; a fixed shuffle does not establish robustness to position. Section~\ref{sec:presentation} examines these cues descriptively using saved responses.

The pinned Hub snapshot, revision \code{609d608...794d}, contains the complete selected set \ifdefined\anonsubmission{}(construction snapshot; repository link withheld for review)\else\citep{preview}\fi, but its manifest is explicitly an accepted-candidate preview with \code{complete:false}. The repository is visible, while file access requires manual approval. Evaluation uses the local completed preparation export, whose Parquet hash matches that snapshot. The distinction is preserved rather than changing a preview flag. Exact hashes and regeneration instructions are supplied in Appendix~\ref{app:repro}.

\section{Evaluation Protocol}\label{sec:protocol}

\subsection{Requests, scoring, and run identity}

Each question is submitted independently with a fixed Turkish prompt requesting one answer letter and no explanation (Appendix~\ref{app:prompts}). Local \code{direct} mode requests \code{think=false} with a 2,048-token output ceiling; \code{thinking} requests \code{think=true} with a 4,096-token ceiling. Both use seed 42 and otherwise retain runtime defaults. The conditions therefore differ in both the reasoning request and token budget: their comparison is not an isolated effect of reasoning.

These ceilings were chosen during calibration after a 42-token ceiling caused some models to exhaust their output budget before producing a visible answer. Direct mode denotes a request, not verified absence of reasoning. Per-response records preserve visible output, reported token counts, truncation, and whether a separate thinking channel or closing reasoning tag was observed. Such flags are backend-dependent measurements, not a complete view of internal computation.

Strict scorer version 2 discards any prefix through the last \code{</think>} tag and unwraps a whole-answer \code{\textbackslash boxed\{X\}} form. It then accepts a valid letter in specified wrappers, normalized exact option text, or a consistent label plus exact option text. Other output is unparseable. This operational definition is deliberately reproducible, although tag handling is a parsing heuristic rather than validation of well-formed reasoning markup. Raw responses are retained, and saved indices and correctness labels are checked against this implementation when generating the paper's tables.

Accuracy is agreement with the source key over all questions; an unparseable response counts as incorrect. Valid-answer rate and invalid/truncated counts distinguish response failures from valid selections that disagree with the key. A truncated response can still score if it contains a parseable answer. Runs with no parseable answers are marked unranked rather than interpreted as zero knowledge. A separate legacy scorer reproduces earlier label heuristics and threshold-free embedding fallback; its scores are diagnostic only.

Benchmark hash, prompt, mode, scorer version, generation settings, runtime, and model identity are retained in run records. The pilot uses a shared comparison hash and separately records each model digest. The full runner includes model identity in each run hash, so the analysis checks common conditions directly rather than equating hashes across models. Only complete runs with compatible conditions are compared; local and API results are displayed separately without a joint rank. Deterministic data export and saved settings do not guarantee bit-identical inference.

\subsection{Calibration and statistical interpretation}\label{sec:pilot}

The fixed 100-question calibration set was sampled from an earlier 7,634-question construction snapshot covering 37 sections (Hub revision \code{cfefc193...c277}). Selection sorts \code{SHA-256(seed + NUL + soru\_id)} with seed 42 and retains the first 100 identifiers. The resulting sample covers 34 sections, and all 100 rows remain identical in the completed benchmark. It is not a random sample of the final 12,000 questions or its 58-section composition. It was used for protocol development and mode selection, so it is a calibration set rather than an untouched validation set.

The pilot covers 25 local tags from eight open-weight families on a Mac mini (M4, 32 GB), Ollama 0.33.2, one request at a time on 1--2 September 2026. Runtime-reported sizes span 0.75--36B parameters. All direct runs completed; storage exhaustion stopped thinking after ten models. That subset reflects execution order and storage, not representative sampling. Appendix~\ref{app:pilot} supplies quantizations and the retained guardrail classifier's task-format failure.

This work reports 95\% Wilson intervals for individual accuracy proportions and paired Wald intervals with exact McNemar tests for changes on the same questions \citep{statistics}. Intervals use an item-independence approximation. Scores on the fixed corpus are observed quantities; these intervals express item-sampling uncertainty under that approximation, not uncertainty from repeated generation, key errors, or dataset construction. Shared donors and related topics may introduce dependence. Pilot mode tests are exploratory, with ten comparisons and a Bonferroni threshold of 0.005. Overlapping marginal intervals are not used as a test of differences between models. No confirmatory ranking inference is claimed.

\subsection{Pilot findings}

Direct-mode accuracy ranges from 8\% to 65\% among models with parseable answers; the guardrail classifier produces none. For the 22 ranked models without observed reasoning, legacy accuracy exceeds strict accuracy by 2--13 points (mean 5.6). A wrong bare letter can reach the legacy embedding fallback and select another option, including the gold. That gold-aware control flow makes the legacy score unsuitable as an alternative knowledge estimate.

Thinking improves accuracy in seven of ten completed pairs, but none passes the Bonferroni threshold. Its budgets differ, and mean request time grows by factors of 2--627. Some direct requests also produce substantial generation or recorded reasoning. These findings motivate strict answer extraction and explicit budget reporting, rather than treating a bare-letter request as equal computation. Appendix~\ref{app:pilot} gives the complete pilot, timing and truncation results.

\section{Complete Evaluations}\label{sec:full}

\subsection{Setup and results}

Four local models from the pilot completed 12,000-question runs on the same Mac mini under Ollama 0.34.0 on 10--11 September 2026: \code{gemma4:31b-mlx}, \code{gemma4:26b-mlx}, \code{gemma4:e4b-mlx}, and \code{qwen3.5:4b-mlx}, all NVFP4 builds. These are the first local runs to complete, not a random sample of models. Their runtime-reported total parameter counts are 31.7B, 26.2B, 8.1B, and 4.5B, respectively; tag sizes need not equal total parameters. The 26B model uses the selector's exact digest.

The fifth run queries the API identifier \code{gpt-6-astra} through OpenAI on 10 September with \code{reasoning\_effort=low}, seed 42, and \code{max\_completion\_tokens=2048}. One request used the standard tier and 11,999 used two Batch jobs. The served-model string remained unchanged, but it is not a weight digest. These are the recorded request settings; they do not establish that other effort levels were unavailable. The API backend, reasoning control, and sampling defaults differ from the local setup.

\begin{table*}[t]
\centering
\small
\setlength{\tabcolsep}{4pt}

\begin{tabular}{@{}lrrcrrr@{}}\toprule
Model & Correct & Accuracy & 95\% CI & Answer & Tokens & Reas.\\\midrule
\makeatletter\code{gemma4:31b} & 7,637 & 63.64 & [62.78, 64.50] & 99.85 & 1.1 & 0.0\\
\code{gemma4:26b}$^{\dagger}$ & 6,901 & 57.51 & [56.62, 58.39] & 99.81 & 1.1 & 0.0\\
\code{gemma4:e4b} & 5,330 & 44.42 & [43.53, 45.31] & 99.94 & 1.0 & 0.0\\
\code{qwen3.5:4b} & 4,181 & 34.84 & [33.99, 35.70] & 99.83 & 1.2 & 0.0\\
\midrule
\code{gpt-6-astra} & 9,766 & 81.38 & [80.68, 82.07] & 99.91 & 27.7 & 46.1\\
 \bottomrule\makeatother
\end{tabular}
\caption{Complete runs ($n=12,000$). Accuracy and answer rate are percentages; intervals are 95\% Wilson intervals. Tokens are mean reported generated tokens per question, including reported reasoning. Reas. is the percentage of responses with observed/reported reasoning. The first four rows are local NVFP4 models; the final row uses an API. Local tags omit \code{-mlx}. $^{\dagger}$Selector weights.}
\label{tab:full-run}
\end{table*}

Strict accuracy ranges from 34.84\% to 81.38\% (Table~\ref{tab:full-run}). Section-macro and item-micro accuracy differ by less than 0.01 percentage points because section sizes are nearly equal. Valid-answer rates are 99.81--99.94\%, with 7--23 invalid responses per model and one truncated response, from \code{qwen3.5:4b-mlx}. The selector scores 57.51\%; without alternative construction models, its participation in selection cannot be assigned a positive or negative effect.

The API run reports reasoning on 5,529 questions (46.1\%); 251,777 of 332,984 output tokens are reported as reasoning tokens. Its accuracy is 96.2\% when reasoning is not reported and 64.1\% when it is. These subsets are selected by the model, so the association identifies neither the causal effect of reasoning nor an independent estimate of question difficulty. Local models average 1.0--1.2 generated tokens per question with no observed reasoning.

The full-run timing and saved-rate cost estimates are documented in Appendix~\ref{app:repro}; they distinguish accumulated local request time from API completion span.

The four local models retain their pilot order, and each full score falls within its pilot interval. However, the pilot is nested in the full benchmark and the runtime changed; this is a consistency check, not independent validation. Removing all 100 calibration questions gives a separate 11,900-question sensitivity analysis (Appendix~\ref{app:repro}). Section scores are reported in Appendix~\ref{app:sections}; their interpretation is limited by roughly 207 questions per section and heterogeneous subject labels.

\subsection{Paired five-versus-ten comparison}\label{sec:five-ten}

The source's mmlu split contains the 6,200 five-choice TR-MMLU questions across 62 sections. Of these, 981 occur in \dataset{}. The analysis reconstructs their original options from provenance and verifies exact equality of stem text, original option text, and source key. The original five-choice order is retained by TR-MMLU, whereas \dataset{} shuffles all ten options. Thus, additions, option order, displayed labels, and prompt length change together. This design measures the implemented augmentation, not the isolated causal effect of adding five options.

The evaluation of \code{gpt-6-astra} on TR-MMLU was conducted on 11 September with the same recorded model identifier, request parameters, and scorer. The prompt changes its letter range from A--J to A--E. One request used the standard tier and 6,199 used a Batch job; two responses were invalid. The five-choice run's estimated token charge is US\$7.16 under its saved rates. Different dates and provider-side nondeterminism remain possible influences despite matching recorded settings.

\begin{table*}[t]
\centering
\footnotesize
\setlength{\tabcolsep}{3pt}

\begin{tabular}{@{}lrrrcrrr@{}}\toprule
Questions & $k$ & $n$ & Accuracy & 95\% CI & Guess-adjusted & Reas. & Tokens\\\midrule
\makeatletter TR-MMLU, all questions & 5 & 6,200 & 93.35 & [92.71, 93.95] & 91.69 & 28.1 & 18.1\\
Turkish MMLU Pro, all questions & 10 & 12,000 & 81.38 & [80.68, 82.07] & 79.31 & 46.1 & 27.7\\
\midrule
Shared questions, original options & 5 & 981 & 93.68 & [91.98, 95.04] & 92.10 & 29.9 & 19.2\\
Shared questions, with borrowed options & 10 & 981 & 83.08 & [80.60, 85.29] & 81.20 & 48.2 & 28.3\\
 \bottomrule\makeatother
\end{tabular}
\caption{Five- and ten-choice results for \code{gpt-6-astra}. Only the 981 shared questions form a paired comparison. All accuracy columns are percentages. Guess-adjusted is $(a-1/k)/(1-1/k)$, with accuracy $a$ on a 0--1 scale and $k$ options; it is a reference transformation, not a validated knowledge estimate. Reas. is reported reasoning prevalence; tokens are mean generated tokens per question.}
\label{tab:five-ten}
\end{table*}

On shared questions, accuracy falls from 919/981 (93.68\%) to 815/981 (83.08\%), a paired change of $-10.60$ percentage points (95\% Wald interval $[-12.74,-8.46]$; exact McNemar $p=5.27\times10^{-23}$). There are 804 questions correct in both conditions, 115 correct only with five choices, 11 correct only with ten, and 51 correct in neither. Of the 115 losses, 102 select borrowed options, 11 select original distractors, and two are invalid. Reported reasoning prevalence rises from 29.9\% to 48.2\% on these questions.

The unpaired full-corpus scores reflect different item compositions. Guess-adjusted columns in Table~\ref{tab:five-ten} are reference transformations, not corrections for ambiguity or presentation effects; Appendix~\ref{app:repro} gives a stylized guessing calculation.

\subsection{Borrowed options and negative stems}\label{sec:analysis}

Across the five full runs, 56.8--70.9\% of valid responses that disagree with the key select a borrowed option. There are five borrowed and four original non-key options per question; evenly distributing such responses would give a borrowed share of $5/9=55.6\%$. This is a descriptive reference, not a realistic behavioral null or proof that borrowed options function as valid distractors. A model can select one because it is misleading or because it is defensible.

Negative stems make the latter mechanism particularly clear. For a question asking which option does \emph{not} belong to a category, a borrowed option outside that category can satisfy the stem alongside the source key. The selector instruction warns against this, but forced selection provides no way to reject an unsuitable candidate pool. The analysis investigates the pattern with a keyword rule applied to the question sentence, including Turkish negation and negative-aorist forms (Appendix~\ref{app:negative}). The rule flags 3,118/12,000 \dataset{} questions and 1,683/6,200 TR-MMLU questions. Section~\ref{sec:audit} compares the rule with the audit records; these are not an independent gold standard, and \emph{unflagged} does not mean confirmed positive.

\begin{table*}[t]
\centering
\small

\begin{tabular}{@{}llrrr@{}}\toprule
Model & Benchmark & Unflagged & Flagged & Gap\\\midrule
\makeatletter\code{gpt-6-astra} & TR-MMLU & 93.7 & 92.5 & +1.2\\
\code{gemma4:31b-mlx} & Turkish MMLU Pro & 71.7 & 40.6 & +31.2\\
\code{gemma4:26b-mlx}$^{\dagger}$ & Turkish MMLU Pro & 65.8 & 33.8 & +32.1\\
\code{gemma4:e4b-mlx} & Turkish MMLU Pro & 50.7 & 26.5 & +24.3\\
\code{qwen3.5:4b-mlx} & Turkish MMLU Pro & 40.3 & 19.4 & +20.9\\
\code{gpt-6-astra} & Turkish MMLU Pro & 88.0 & 62.6 & +25.3\\
 \bottomrule\makeatother
\end{tabular}
\caption{Source-key accuracy (\%) by the approximate negative-stem rule. Unflagged/flagged sample sizes are 8,882/3,118 for \dataset{} and 4,517/1,683 for TR-MMLU. Gap is unflagged minus flagged, in percentage points. $^{\dagger}$Selector weights.}
\label{tab:polarity}
\end{table*}

All five models score 20.9--32.1 points lower on flagged stems in \dataset{} (Table~\ref{tab:polarity}). The corresponding gap for the API model on TR-MMLU is 1.2 points. These full-corpus contrasts can reflect subject composition as well as wording; the paired subset provides a more focused diagnostic.

\begin{table*}[t]
\centering
\small
\setlength{\tabcolsep}{4pt}

\begin{tabular}{@{}lrrrrcr@{}}\toprule
Stem group & $n$ & Five & Ten & Change & 95\% CI & Lost/gained\\\midrule
\makeatletter Flagged negative & 250 & 88.4 & 64.0 & -24.4 & [-30.3, -18.5] & 66/5\\
Unflagged & 731 & 95.5 & 89.6 & -5.9 & [-7.8, -3.9] & 49/6\\
 \bottomrule\makeatother
\end{tabular}
\caption{Post hoc paired changes on the 981 shared questions, split by the same rule. Accuracy is percent; change is ten minus five choices in percentage points, with a paired 95\% Wald interval. Lost/gained counts refer to source-key correctness.}
\label{tab:paired-polarity}
\end{table*}

Among the 250 flagged shared questions, accuracy decreases from 88.4\% to 64.0\% ($-24.4$ points), versus $-5.9$ points among 731 unflagged questions (Table~\ref{tab:paired-polarity}). Flagged questions account for 61 of the 104 net losses. Of their 66 previously correct answers that become incorrect, 64 select borrowed options; 38 of the 49 corresponding losses on unflagged questions do so. This response pattern alone does not identify ambiguous items. The separate audit examines recorded answer validity on a corpus sample, rather than adjudicating all 981 paired items. Unflagged losses likewise do not establish valid distractor difficulty.

\subsection{Presentation and source-overlap diagnostics}\label{sec:presentation}

Post hoc diagnostics describe position and length associations using saved responses (Table~\ref{tab:presentation}). Invalid responses remain incorrect in position-specific accuracy; entropy uses valid responses. Error primacy compares A selections on errors with a uniform reference over each item's nine non-key positions. Appendix~\ref{app:diagnostics} defines all quantities. These are associations across different questions, not within-item interventions.

\begin{table*}[t]
\centering
\small
\setlength{\tabcolsep}{4pt}

\begin{tabular}{@{}lrrrrr@{}}\toprule
Model & Range & Slope & Entropy & A excess & Length rank\\\midrule
\makeatletter\code{gemma4:31b} & 58.4--67.7 & -0.96 & 0.996 & +3.0 & 0.525\\
\code{gemma4:26b} & 52.9--60.7 & -0.68 & 0.997 & +0.9 & 0.523\\
\code{gemma4:e4b} & 42.3--50.6 & -0.19 & 0.994 & +0.1 & 0.536\\
\code{qwen3.5:4b} & 17.4--56.1 & -4.57 & 0.886 & +10.9 & 0.515\\
\code{gpt-6-astra} & 79.4--85.1 & -0.45 & 0.999 & +4.9 & 0.528\\
 \bottomrule\makeatother
\end{tabular}
\caption{Descriptive diagnostics on the five complete ten-choice runs. Range: minimum--maximum gold-position accuracy (\%). Slope: item-level linear accuracy slope in percentage points per later gold position. Entropy: response entropy divided by $\log 10$. A excess: error primacy in percentage points. Length rank: mean selected-option length midrank among the nine non-key options, on valid errors (0 shortest, 1 longest; uniform reference 0.5).}
\label{tab:presentation}
\end{table*}

The largest position association occurs for \code{qwen3.5:4b-mlx}: accuracy ranges from 17.4\% to 56.1\%, with a slope of $-4.57$ points per later gold position and an A-error excess of 10.9 points. The API model has a narrower 79.4--85.1\% range and near-uniform aggregate response entropy (0.999), yet its A-error excess is 4.9 points. High aggregate entropy therefore does not establish absence of position association. Since different questions occupy different gold positions, these statistics do not measure the effect of moving a fixed answer or partition the paired augmentation loss.

Within a question, an original option is longer than a borrowed option in 58.84\% of original--borrowed comparisons, counting ties as one half. This indicates a modest length cue to origin. On valid errors, mean selected length ranks are 0.515--0.536, slightly above the uniform non-key reference. In an exploratory subset of 151 shared questions whose borrowed mean length is within 10\% of the original mean, accuracy still decreases from 141/151 (93.38\%) to 127/151 (84.11\%): 15 losses and one gain, a change of $-9.27$ points (paired Wald interval $[-14.25,-4.29]$). This text-based restriction is a sensitivity check, not a length intervention: it neither matches full length distributions nor controls total prompt length, content, or option order.

Provenance also documents item coupling. Of 33,617 distinct donors, 9,068 supply options to multiple recipients; 9,601/12,000 selected questions share at least one donor with another recipient. Also, 11,986/60,000 additions come from questions that are themselves in the selected set, affecting 5,518 recipients. There are 15,682 additions across original source-split boundaries. Normalized-stem comparison finds no duplicate stems within the selected set, but 94 selected stems also occur in another raw source split; deduplication retains one owner rather than removing upstream copies. These are construction dependencies and source overlaps, not evidence that a model encountered the questions during training. They reinforce the need to avoid treating source split names or donor-linked records as independent evaluation partitions.

\begin{table*}[t]
\centering
\small

\begin{tabular}{@{}lrrrr@{}}\toprule
Answer-set classification & R1 $n$ & R1 W\% & R2 $n$ & R2 W\%\\\midrule
\makeatletter Unique source key & 152 & 80.8 & 119 & 68.1\\
Multiple answers & 47 & 18.8 & 31 & 14.9\\
Single alternative & 1 & 0.5 & 25 & 9.0\\
Unresolved & 0 & 0.0 & 25 & 8.0\\
\midrule
Key + borrowed answer & 43 & 15.6 & 15 & 6.3\\
 \bottomrule\makeatother
\end{tabular}
\caption{Reconciled audit records on the same 200 sampled questions. $n$ is the observed count; W\% weights each question by its inverse inclusion probability and uses the full 12,000 total weight, including unresolved items. The first four rows partition each record set. The final row is a subset of multiple-answer judgments. These are descriptive record summaries, not certified corpus-validity rates.}
\label{tab:audit}
\end{table*}

\subsection{Completed audit and reconciled records}\label{sec:audit}

The audit samples 200 questions from all 58 sections and 116 section-by-flag strata, including 101 flagged and 99 unflagged items. Sampling precedes annotation; seed-42 hash ordering and fixed inclusion weights are retained. Twelve separate calibration items are excluded. Each sampled item has two sets of reviewer records, denoted R1 and R2, and five borrowed-option judgments per set. The author reports the unchanged rubric, calibration before labelling, full 200-item coverage and independence until reconciliation; both reviewers are native Turkish speakers with university degrees, without per-discipline expertise. Reviewer AI use is not established, and R2's second-pass fields are script-generated, so the records are human-checked but unverified (Appendix~\ref{app:audit}).

A separate 1,054-entry correction log resolves exported labels while preserving original records. Corrections explicitly use the other reviewer's answers, so the final records are reconciled judgments, not independent repeated measurements. For 46 R2 questions the log supplies revised answer sets; 25 are unresolved. Dependent status labels are derived from those sets and kept separate from recorded judgments. Blank rubric fields remain missing. Appendix~\ref{app:audit} details the procedure and limitations.

R1 records 47 multiple-answer items; R2 records 31 and leaves 25 unresolved (Table~\ref{tab:audit}). The source key and a borrowed option are both marked defensible in 43 and 15 items, respectively. These observations support concern that borrowing can introduce competing answers, but cross-reviewer dependence and unresolved cases prevent a pooled validity estimate. The original polarity fields agree with the keyword rule on 187/200 and 188/200 items; their weighted counts appear in Appendix~\ref{app:audit}. This checks recorded polarity consistency without establishing expert-gold precision or recall. All model scores retain the original frozen keys; no response is rescored against the audit.

\section{Discussion}

The construction establishes text preservation and provenance, and the saved-response comparison establishes lower source-key agreement after augmentation. The completed audit provides human-checked records of multiple defensible answers, with disagreement and unresolved items retained. Together, these results support a validity failure mode worth investigating. They do not establish that ten-choice scores measure Turkish knowledge better, or separate ambiguity from presentation effects. A larger, documented domain-expert adjudication and controlled reevaluation are needed for a validated replacement benchmark.

A revised construction should permit fewer additions or reject an item when five verified non-answer options cannot be found. Negative questions require each addition to fail the requested exception. Lexical filtering alone will miss problems and remove valid questions. Revised keys/options require a new artifact version and new evaluations; the current scores cannot be transferred. Reporting flagged strata makes the observed weakness visible without certifying unflagged items.

\section{Conclusion}

\dataset{} makes option borrowing traceable across 12,000 Turkish-source questions. The paired analysis finds a 10.6-point decrease in source-key agreement, concentrated on flagged negative stems. The completed audit records competing answers and unresolved judgments, so lower scores do not establish better assessment. A validated successor needs semantic rejection at construction, expert adjudication and controlled reevaluation.

\label{main-content-end}
\clearpage

\section*{Limitations}\label{sec:limitations}

The audit is completed, but its evidential scope is limited. Unverified reviewer tool use and cross-reviewer reconciliation preclude treating the final records as independent expert ratings. The reported protocol order is the author's account rather than a property established by the retained records, and domain-specific qualifications are not claimed. Twenty-five R2 answer sets remain unresolved, and other missing fields are not imputed. Inverse-inclusion weighting corrects sampling allocation, not annotation error. Reported audit percentages are descriptive; no confidence interval or corpus certification is inferred from them.

Several design limits constrain the paired result. It covers one model, one fixed presentation per condition, and a selected overlap of 981 questions rather than all source items. Original options are reordered and relabelled during export, and requests occur on different dates. These controls are specified rather than left open. A four-condition reevaluation is designed: a five-choice baseline in original A--E order; ten choices with the original options fixed and borrowed options appended F--J, isolating option count from reordering; ten choices interleaved under frozen seeds, preserving relative order while changing labels; and unrestricted frozen permutations including the evaluated export, re-run concurrently rather than compared against the historical run. Repeated permutations require analysis clustered by question rather than treated as independent items, with seeds fixed before inference and no post hoc seed selection. Extending the paired design to several models and to an alternate selector would separate construction-model effects from augmentation. Independent item adjudication by domain specialists, without cross-reviewer reconciliation, should precede any claim about increased valid difficulty. None of these has been executed, and no result is reported or implied for them. The subgroup analysis was motivated after inspecting responses and remains exploratory.

The calibration set was drawn from an earlier, incomplete section roster, reused during protocol development, and retained in the full benchmark. Runtime defaults, quantization, a single prompt, and one generation per item limit generalization; seed recording does not quantify repeat-run variance. The ten completed thinking runs form an incomplete subset and differ from direct runs in both requested reasoning and output ceiling. The full results cover four selected local builds and one API identifier. API-reported model names and reasoning tokens offer weaker control than pinned weights, and a common direct-mode label does not imply equal computation.

External leaderboard and publisher scores do not provide controlled language comparisons. The available TR-MMLU leaderboard uses older scoring and different or unrecorded quantizations, while publisher evaluations differ in data and harness settings. The experimental comparison is therefore restricted to saved runs whose conditions can be inspected. The research dossier retains the external scores as context, without using their rank correlations or score gaps as evidence for augmentation.

The source was accessible before construction and includes earlier training and evaluation splits. Exact stem deduplication does not eliminate paraphrases or possible pretraining exposure. The overlap and donor audit documents source reuse, but no model-training contamination test or independent human difficulty study has been conducted. Near-equal section sizes are an explicit weighting choice; they do not represent the educational or social importance of subjects. These scores concern prompted answer selection, not comprehensive Turkish ability, safety, or suitability for consequential decisions.

The pinned source card has conflicting license labels: machine-readable CC BY-NC-ND 4.0 and prose CC BY-NC 4.0, with additional third-party material noted. This records an unresolved provenance and distribution question, not a legal determination. The curator reports that the underlying material was collected from public, often unattributed sources. That provenance statement does not itself establish item-level redistribution permissions. Dataset documentation follows the provenance and intended-use concerns of datasheets \citep{datasheets}. Researcher verification and anti-training terms are access policies, not a resolution of these source-rights questions.

\section*{Ethical Considerations}\label{sec:ethics}

\ifdefined\anonsubmission\else
\paragraph{Author contributions.} M. Ali Bayram conceived the study, developed the dataset and evaluation software, conducted the experiments, analysed the results, and wrote and revised the manuscript.
\fi

\paragraph{Funding and competing interests.} This work received no specific funding. The author declares no financial or non-financial competing interests.

\paragraph{Ethics and consent.} The work uses an existing question collection, automated model responses and a completed human-checked audit. The author reports recruiting both reviewers through personal contacts. Both participated as unpaid volunteers and received no compensation. They were informed informally that their labels would support a published paper and that anonymized records would be released, and both gave informal verbal consent; no written consent was collected. The work was never submitted to an institutional ethics board, and no approval or exemption is claimed or implied. Reviewer records are reported under R1/R2 identifiers; beyond native Turkish proficiency and a university degree, no further demographic or professional credentials are collected or inferred. Source licensing and third-party content are discussed in the Limitations section.

\paragraph{Data and code availability.} Three artifacts are distinguished. First, the selected-set snapshot is identified by its pinned Hub revision \ifdefined\anonsubmission{}(construction snapshot; repository link withheld for review)\else\citep{preview}\fi. Its metadata marks it an accepted-candidate preview, but its 12,000-question Parquet file is byte-identical to the completed export: \emph{preview} names the repository's release stage, not a smaller or provisional dataset. Second, the completed manifest, evaluation responses and analysis scripts are local artifacts without an established public archive identifier. Third, donor provenance records and the audit packet remain private because they contain source question text. On 12 September 2026 the repository was publicly visible but manually gated, and unauthenticated file requests were denied; visibility therefore does not establish unrestricted download, which requires repository approval. The source permission conflict described in the Limitations section remains unresolved; repository visibility does not resolve permission for a final transformed-data release. The planned release will retain manual access approval for verified researchers, with evaluation-only terms prohibiting use of the test set for model training or fine-tuning and unauthorized redistribution. These measures aim to reduce contamination; they cannot guarantee that previously accessible source questions are absent from training corpora. Peer-review access must preserve reviewer anonymity and must not require reviewers to identify themselves to the author.

\paragraph{Use of generative AI.} A language model selected existing option identifiers during dataset construction as described in Section~\ref{sec:selection}. AI assistance supported manuscript review and revision, literature checks, diagnostic analysis code and audit preparation. Whether the two audit reviewers used AI tools while producing their judgments could not be established: no per-reviewer tool record was kept, and the author cannot confirm their working practice. The returned material includes a model-query script whose execution status is unknown, and second-pass fields that are reproducibly script-generated. Because reviewer tool use cannot be excluded, the records are reported as human-checked but unverified rather than as independent human adjudication. Explicit corrections, missingness and reconciliation are retained rather than treating automated fields as independently validated judgments. The author has reviewed the text, analyses, and references and accepts responsibility for them. Reconciled records of unverified provenance are not represented as independent expert adjudication.

\ifdefined\tmlrsubmission
  \bibliographystyle{tmlr}
\fi
\bibliography{references}
\clearpage
\appendix
\section{Prompts and implementation details}\label{app:prompts}

The evaluation prefix is:
\begin{quote}\small\ttfamily\raggedright
Sana soru ve seçenekleri veriyorum. Sadece hangi seçeneğin sorunun doğru cevabı olduğunu yaz. A--J arasındaki tek bir harfle cevap ver. Lütfen herhangi bir açıklama yapma!
\end{quote}
It is followed by \code{Soru:} (``Question:), the stem, and options labelled \code{A:} through \code{J:}, one per line. In English, the instruction asks for the correct option as a single letter between A and J and requests no explanation. The five-choice condition replaces A--J with A--E. The executable prompt uses an en dash in the range; the full Unicode string is preserved in the run records.

The selector's complete Turkish instruction and dynamic candidate-ID schema are recorded in \code{research/prompts.json} and \code{prepare.py}; an English translation is supplied in \path{research/prompt_translations.json}. It instructs the model to treat the embedded question JSON as data, trust the original key, avoid alternative answers and semantic duplicates, handle negative stems, select five IDs even when candidates are weak, and return only the requested JSON. The saved selector model digest is \code{f0fc7e0a...64879}. The historical code snapshot is retained separately from the current analysis scripts.

\section{Negative-stem rule}\label{app:negative}

The rule normalizes Unicode and lowercases with Turkish dotted and dotless \emph{i} handled explicitly. It takes the last sentence containing a question mark or \emph{hangi} (``which''), falling back to the last sentence. It flags matches for forms beginning with \emph{değil}, \emph{yanlış}, \emph{dışında}, or \emph{olmayan}; the words \emph{yok}, \emph{yoktur}, \emph{olamaz}, \emph{söylenemez}, and \emph{hariç}; or negative-aorist endings \emph{-maz}/\emph{-mez} with the implemented plural and copular suffixes. It excludes exact matches for \emph{namaz}, \emph{yılmaz}, and \emph{taşınmaz}. The executable expression in \code{paper/pilot\_results.py} is authoritative. Attributive negation and ambiguous suffixes can create false positives, and other negative constructions can be missed. The audit comparison in Appendix~\ref{app:audit} reports agreement with the recorded polarity fields; their provenance does not support treating them as an independent gold standard.

The lexical meanings include ``not'' (\emph{değil}, \emph{olmayan}), ``wrong'' (\emph{yanlış}), ``outside/except'' (\emph{dışında}, \emph{hariç}), ``absent'' (\emph{yok}, \emph{yoktur}), ``cannot be'' (\emph{olamaz}) and ``cannot be said'' (\emph{söylenemez}). The excluded forms mean ``ritual prayer'' (\emph{namaz}), ``undaunted'' (\emph{yılmaz}, also a name) and ``immovable/property'' (\emph{taşınmaz}); suffix-shaped spelling alone does not determine polarity.

\section{Diagnostic definitions}\label{app:diagnostics}

\code{paper/review\_diagnostics.py} verifies the inputs pinned by the evaluation snapshot and rescores the full responses before computing Table~\ref{tab:presentation}. The gold-position slope is ordinary least squares on per-item correctness (scaled to percent), with an intercept and the zero-based gold index as predictor. Response entropy is $-\sum_j p_j\log p_j/\log 10$, using valid-response frequencies. Error primacy is $100[E_A-\sum_{i\in\mathcal E}\mathbf{1}(g_i\ne A)/9]/|\mathcal E|$, where $\mathcal E$ contains valid errors, $E_A$ counts selections of A, and $g_i$ is the source key. Length is the number of Unicode code points after trimming display whitespace. Length ranks average tied ranks and exclude the gold when analyzing errors. None of these post hoc metrics is a test of semantic validity or a within-item position intervention.

\section{Audit sampling, corrections and interpretation}\label{app:audit}

The original packet uses seed-42 SHA-256 ordering to allocate 3--4 items per section, split across the heuristic-flag strata, and samples without replacement. All 116 nonempty strata are represented. For item $i$ in stratum $h$, $w_i=N_h/n_h$; the 200 weights sum to 12,000. Table~\ref{tab:audit} reports $100\sum_i w_i\mathbf{1}(y_i=c)/\sum_i w_i$. Unresolved labels are a separate category in the same denominator. The 12 disjoint calibration items do not enter any reported statistic. No sampling-based interval is claimed: sparse strata, annotation uncertainty and cross-reviewer dependence are not captured by a simple binomial interval.

The supplied instructions ask for all defensible A--J answers, stem polarity, clarity, missing context, subject familiarity and references before revealing keys/origins. The second pass asks for source-key support, multiple/no valid answers, semantic duplicates and section fit, plus each borrowed option's validity, relevance, duplicates and incomplete references. These are separate dimensions: an incorrect but unrelated option is not evidence of a plausible distractor. The original instructions are supplied with the reviewer materials. The planned concealment in the HTML is not evidence that the actual first pass was blind; Q159's correction explicitly acknowledges deference to a source key.

The original six CSVs contain 200 first-pass and 200 second-pass question rows plus 1,000 option rows for each of R1 and R2. A separate correction log has 1,054 entries: 33 R2 source-key category corrections, 173 option resolutions, two R1 Q159 entries, 800 explicit missing-field entries, and 46 final R2 answer sets. Of the 173 option resolutions, 72 retain the earlier answer interpretation, 14 retain the distractor interpretation, and 87 retract both and classify the option as defective. The latter two groups retract 101 answer selections across 46 questions. Twenty-one retain an explicitly recorded answer set after subtraction; 25 are unresolved, not verified no-answer items. The subtraction-based sets are not described as fresh expert re-adjudication.

The analysis preserves received records and derives answer-set status separately. A singleton equal to the source key is unique-source-key; another singleton is single-alternative; a set with more than one answer is multiple; an unresolved set yields cannot-verify. Derived status updates affect 42 R2 answer-status and 25 key-support fields. A multiple set may still exclude the source key, so multiple-answer status and key support are distinct. A source-key-plus-borrowed judgment requires both the source key and at least one borrowed position in a resolved set. Missing data never become negative findings. R2's unresolved items carry 963/12,000 total sampling weight (8.0\%); its observed weighted co-answer fraction is 6.3\%, with an unknown-allocation range of 6.3--14.3\% if every unresolved item could qualify. That range describes missingness alone, not statistical uncertainty or ground-truth prevalence.

The corrections cite R1 when reconciling R2, so post-correction agreement would be circular evidence of independent reliability. Neither Cohen's kappa nor expert-gold precision/recall is claimed. R2's original second-pass fields are script-generated rather than judged: re-running the supplied generation script over the returned first-pass answers and the revealed-key page reproduces both R2 second-pass files byte for byte. Its 1,000 option rows are constant, marking every borrowed option a valid, plausible, non-duplicate distractor. These entries carry no information and are never promoted to verified plausibility ratings; no reported quantity depends on them. R1's second-pass option fields vary across all four judgment columns and retain the shipped template's byte signature. The main results therefore emphasize answer-set records and unresolved coverage rather than a pooled option-quality percentage. The artifact itself is unchanged; revised keys/options would require a new version and new evaluations.

\begin{table*}[t]
\centering
\small

\begin{tabular}{@{}lrrrr@{}}\toprule
Flag / recorded polarity & R1 $n$ & R1 weight & R2 $n$ & R2 weight\\\midrule
\makeatletter Flagged / negative & 90 & 2833.0 & 91 & 2864.5\\
Flagged / nonnegative & 11 & 285.0 & 10 & 253.5\\
Unflagged / negative & 2 & 149.0 & 2 & 149.0\\
Unflagged / nonnegative & 97 & 8733.0 & 97 & 8733.0\\
 \bottomrule\makeatother
\end{tabular}
\caption{Keyword flags against recorded audit polarity. Counts and inverse-inclusion-weight sums are reported separately for R1 and R2. The 200 original polarity labels in each record set are unchanged by the correction log. These are recorded-label comparisons, not validated expert-gold confusion matrices.}
\label{tab:audit-polarity}
\end{table*}

\section{Detailed pilot results}\label{app:pilot}

Among models with parseable answers, direct-mode accuracy ranges from 8\% to 65\%. Nine models score between 57\% and 65\%, with broad intervals at this sample size; their observed ordering is descriptive. The guardrail classifier supplies no valid answers. \code{minicpm-v4.6:1b} answers 28 of 100 questions in a parseable form, with 8\% overall accuracy and 28.6\% accuracy on those 28 answers. Answered-only accuracy conditions on a model-selected subset and cannot replace the overall score.

Generation counts vary substantially within direct mode. The \code{muse-glimmer:30b-mlx} run averages 266 generated tokens per question despite returning short visible answers and no recorded reasoning trace. This indicates substantial generation not represented by the visible answer; its semantic content cannot be inferred from the count alone. \code{lfm2.5:8b} has observed reasoning in 84 responses, with 16 reaching the 2,048-token ceiling. A request for a bare letter therefore does not equalize computation.

For the 22 ranked models without observed reasoning, legacy accuracy exceeds strict accuracy by 2--13 percentage points (mean 5.6). Inspection of the legacy code explains why this is not simply recovered answer coverage: a wrong bare letter can reach an embedding fallback that selects another option, including the gold. Its apparent success therefore depends partly on the gold-aware control flow. This diagnostic must not be interpreted as a valid alternative estimate of knowledge.

Thinking-mode accuracy is higher for seven of ten completed pairs, by 3--11 points, and lower for three. Only \code{gemma4:e4b-mlx} reaches unadjusted $p<0.05$ ($+11$ points; $p=0.035$); none meets the Bonferroni threshold. Generation time rises by factors of 2--627. For \code{qwen3.5:0.8b-mlx}, \code{qwen3.5:2b-mlx}, and \code{granite4.2:3b}, 48--80 thinking responses reach the ceiling without an answer. The project's dated analysis contract records selection of direct mode for full evaluation on 10 September while the first full run was in progress. This is an internal decision record, not an independently registered protocol. All models use that mode; the protocol does not select each model's better pilot condition.

Tables~\ref{tab:pilot-direct} and~\ref{tab:pilot-modes} report the complete pilot and its ten completed mode pairs. Displayed ranks merely order observed scores. Token counts include generation reported by the runtime, and timing is elapsed request time per question. The \code{-mlx} suffix is omitted from local MLX model names to fit the tables; exact tags, digests, and quantizations are retained in the machine-readable results. Selector weights are marked with a dagger.

\begin{table*}[t]
\centering
\footnotesize
\setlength{\tabcolsep}{2.5pt}

\begin{tabular}{@{}rlrlrcrrrrr@{}}\toprule
\# & Model & Params & Quant. & Acc. & 95\% CI & Ans. & Legacy & Tokens & Reas. & s/q\\\midrule
\makeatletter 1 & \code{muse-glimmer:30b} & 32.3 & nvfp4 & 65 & [55, 74] & 100 & 67 & 266 & 0 & 13.65\\
2 & \code{gemma4:31b} & 31.7 & nvfp4 & 63 & [53, 72] & 100 & 66 & 1 & 0 & 1.64\\
3 & \code{qwen3.8:27b} & 27.8 & nvfp4 & 63 & [53, 72] & 100 & 66 & 1 & 0 & 1.53\\
4 & \code{gemma4:26b}$^{\dagger}$ & 26.2 & nvfp4 & 60 & [50, 69] & 99 & 63 & 1 & 0 & 0.39\\
5 & \code{qwen3.6:27b} & 27.8 & nvfp4 & 60 & [50, 69] & 100 & 65 & 1 & 0 & 1.48\\
6 & \code{qwen3.6:35b} & 36.0 & nvfp4 & 59 & [49, 68] & 100 & 64 & 1 & 0 & 0.53\\
7 & \code{qwen3.5:27b} & 27.4 & nvfp4 & 58 & [48, 67] & 100 & 62 & 1 & 0 & 1.51\\
8 & \code{qwen3.5:35b} & 35.1 & nvfp4 & 57 & [47, 66] & 100 & 61 & 1 & 0 & 2.17\\
9 & \code{gemma4:12b} & 12.4 & nvfp4 & 55 & [45, 64] & 100 & 57 & 1 & 0 & 0.73\\
10 & \code{ornith-1.5:35b} & 35.5 & Q4\_K\_M & 52 & [42, 62] & 96 & 55 & 6 & 0 & 0.54\\
11 & \code{qwen3.5:9b} & 9.4 & nvfp4 & 48 & [38, 58] & 100 & 52 & 1 & 0 & 0.49\\
12 & \code{ornith-1.5:9b} & 9.0 & Q4\_K\_M & 45 & [36, 55] & 100 & 49 & 2 & 0 & 0.56\\
13 & \code{gemma4:e4b} & 8.1 & nvfp4 & 42 & [33, 52] & 100 & 46 & 1 & 0 & 0.25\\
14 & \code{granite4.2:30b} & 29.3 & Q4\_K\_M & 42 & [33, 52] & 100 & 49 & 2 & 0 & 1.49\\
15 & \code{ornith:9b} & 9.0 & Q4\_K\_M & 41 & [32, 51] & 100 & 49 & 2 & 0 & 0.62\\
16 & \code{qwen3.5:4b} & 4.5 & nvfp4 & 40 & [31, 50] & 100 & 49 & 1 & 0 & 0.29\\
17 & \code{granite4.2:8b} & 8.8 & Q4\_K\_M & 31 & [23, 41] & 100 & 39 & 2 & 0 & 0.51\\
18 & \code{nemotron-3.5-lightning:30b} & 32.9 & nvfp4 & 31 & [23, 41] & 99 & 40 & 1 & 0 & 0.81\\
19 & \code{gemma4:e2b} & 5.2 & nvfp4 & 30 & [22, 40] & 100 & 37 & 1 & 0 & 0.15\\
20 & \code{lfm2.5:8b} & 8.5 & Q4\_K\_M & 19 & [13, 28] & 71 & -- & 1151 & 84 & 7.94\\
21 & \code{qwen3.5:2b} & 2.2 & mxfp8 & 16 & [10, 24] & 100 & 25 & 2 & 0 & 0.16\\
22 & \code{granite4.2:3b} & 3.7 & Q4\_K\_M & 13 & [8, 21] & 94 & -- & 41 & 2 & 0.66\\
23 & \code{qwen3.5:0.8b} & 0.85 & mxfp8 & 13 & [8, 21] & 100 & 20 & 1 & 0 & 0.09\\
24 & \code{minicpm-v4.6:1b} & 0.75 & Q4\_K\_M & 8 & [4, 15] & 28 & 21 & 186 & 0 & 0.98\\
-- & \code{granite4.1-guardian:8b} & 8.4 & Q6\_K & \multicolumn{1}{c}{unranked} & -- & 0 & -- & 9 & 0 & 0.98\\
 \bottomrule\makeatother
\end{tabular}
\caption{Direct-mode pilot, 100 questions per model. Params: runtime-reported billions of parameters. Acc.: source-key accuracy (\%) with Wilson interval. Ans.: valid-answer rate (\%). Legacy: diagnostic accuracy, withheld for models with observed reasoning or no valid answers. Tokens: mean generated tokens. Reas.: observed reasoning count. s/q: seconds per question.}
\label{tab:pilot-direct}
\end{table*}

\begin{table*}[t]
\centering
\footnotesize
\setlength{\tabcolsep}{3pt}

\begin{tabular}{@{}lrrrcrcrrr@{}}\toprule
Model & Direct & Thinking & $\Delta$ & 95\% CI & $p$ & D/T & Answer (\%) & Trunc. & Time\\\midrule
\makeatletter\code{gemma4:e4b} & 42 & 53 & +11 & [1.9, 20.1] & 0.035 & 6/17 & 100 & 0 & 42$\times$\\
\code{granite4.2:3b} & 13 & 22 & +9 & [0.2, 17.8] & 0.078 & 6/15 & 52 & 48 & 108$\times$\\
\code{ornith-1.5:35b} & 52 & 59 & +7 & [-1.0, 15.0] & 0.143 & 5/12 & 89 & 10 & 59$\times$\\
\code{qwen3.6:35b} & 59 & 66 & +7 & [0.1, 13.9] & 0.092 & 3/10 & 96 & 4 & 85$\times$\\
\code{qwen3.5:4b} & 40 & 46 & +6 & [-3.9, 15.9] & 0.327 & 10/16 & 66 & 34 & 329$\times$\\
\code{gemma4:26b}$^{\dagger}$ & 60 & 64 & +4 & [-3.8, 11.8] & 0.455 & 6/10 & 85 & 15 & 126$\times$\\
\code{lfm2.5:8b} & 19 & 22 & +3 & [-4.6, 10.6] & 0.607 & 6/9 & 83 & 5 & 2$\times$\\
\code{gemma4:e2b} & 30 & 29 & -1 & [-8.1, 6.1] & 1.000 & 7/6 & 76 & 0 & 16$\times$\\
\code{qwen3.5:2b} & 16 & 15 & -1 & [-8.6, 6.6] & 1.000 & 8/7 & 20 & 80 & 627$\times$\\
\code{qwen3.5:0.8b} & 13 & 10 & -3 & [-11.5, 5.5] & 0.648 & 11/8 & 27 & 73 & 499$\times$\\
 \bottomrule\makeatother
\end{tabular}
\caption{Paired thinking versus direct pilot results. $\Delta$ is thinking minus direct in percentage points, with a paired Wald interval. Exact McNemar $p$-values are unadjusted. D/T counts direct-only/thinking-only correct answers. Answer and truncation refer to thinking mode. Time is the ratio of mean request durations. The reasoning request and token ceiling both differ between conditions.}
\label{tab:pilot-modes}
\end{table*}

\section{Full-benchmark accuracy by section}\label{app:sections}

Table~\ref{tab:full-sections} reports the five complete runs by source section. These descriptive scores retain the source keys and do not constitute independently validated subject-level measurements. Appendix~\ref{app:glosses} supplies English glosses of the inherited labels.

\begin{table*}[t]
\centering
\footnotesize
\setlength{\tabcolsep}{2pt}

\begingroup
\renewcommand{\code}[1]{\texttt{\scriptsize #1}}
\begin{tabular}{@{}lrrrrrr@{}}\toprule
Section & \code{gemma4:31b} & \code{gemma4:26b} & \code{gemma4:e4b} & \code{qwen3.5:4b} & \code{gpt-6-astra} & Mean\\
\midrule
Ehliyet & 82.6 & 75.8 & 58.9 & 55.6 & 92.8 & 73.1\\
Dini Bilgiler & 76.3 & 71.0 & 53.1 & 40.6 & 89.9 & 66.2\\
Kim 500 Milyar İster & 79.2 & 72.9 & 48.3 & 30.9 & 96.1 & 65.5\\
Turizm ve Otel İşletmeciliği & 71.4 & 67.0 & 51.9 & 45.1 & 87.9 & 64.7\\
TUS & 78.3 & 72.9 & 42.0 & 35.3 & 89.4 & 63.6\\
İşletme Yönetimi & 67.6 & 63.3 & 52.7 & 39.6 & 87.4 & 62.1\\
Radyo ve Televizyon Programcılığı & 71.5 & 65.7 & 51.2 & 37.7 & 84.1 & 62.0\\
Sosyal Hizmet & 69.6 & 59.9 & 52.2 & 40.6 & 83.1 & 61.1\\
Tıbbi Dökümantasyon ve Sekreterlik & 70.5 & 63.3 & 53.6 & 31.9 & 83.1 & 60.5\\
İktisat & 67.1 & 63.3 & 48.3 & 39.1 & 84.5 & 60.5\\
Siyer & 72.9 & 66.2 & 36.7 & 34.3 & 89.9 & 60.0\\
Medya ve İletişim & 67.6 & 64.3 & 49.3 & 37.2 & 80.7 & 59.8\\
Sağlık Yönetimi & 67.1 & 60.4 & 48.3 & 41.1 & 80.2 & 59.4\\
Sosyal Hizmetler & 68.6 & 58.9 & 48.8 & 39.1 & 81.6 & 59.4\\
Sosyoloji & 66.2 & 61.4 & 51.2 & 36.7 & 81.6 & 59.4\\
Sağlık Kurumları İşletmeciliği & 65.2 & 62.3 & 48.3 & 41.5 & 79.2 & 59.3\\
Elektrik Enerjisi Üretim,İletim ve Dağıtımı & 69.1 & 59.9 & 47.8 & 36.7 & 82.6 & 59.2\\
Menkul Kıymetler ve Sermaye Piyasası & 69.6 & 60.4 & 50.2 & 35.7 & 80.2 & 59.2\\
Kamu Yönetimi & 64.7 & 60.9 & 47.3 & 42.5 & 79.7 & 59.0\\
Bankacılık ve Sigortacılık & 63.8 & 63.3 & 48.8 & 35.3 & 83.6 & 58.9\\
Yerel Yönetimler & 67.6 & 57.0 & 42.0 & 42.0 & 86.0 & 58.9\\
Kültürel Miras ve Turizm & 67.6 & 58.0 & 45.9 & 37.2 & 84.5 & 58.6\\
İlahiyat & 64.3 & 59.9 & 49.3 & 38.2 & 80.2 & 58.4\\
Büro Yönetimi ve Yönetici Asistanlığı & 64.7 & 58.5 & 49.3 & 39.1 & 79.7 & 58.3\\
Tarım & 65.7 & 57.5 & 43.5 & 37.2 & 86.0 & 58.0\\
Turizm ve Seyehat Hizmetleri & 63.3 & 56.5 & 44.4 & 39.1 & 86.5 & 58.0\\
Özel Koruma ve Güvenlik & 63.8 & 59.9 & 44.0 & 37.2 & 82.6 & 57.5\\
Çağrı Merkezi Hizmetleri & 60.4 & 61.4 & 46.9 & 33.3 & 83.1 & 57.0\\
Parakende Satış ve Mağaza Yöneticiliği & 61.4 & 56.5 & 48.3 & 40.1 & 76.3 & 56.5\\
Yönetim Bİlişim Sistemleri & 65.7 & 56.5 & 48.3 & 34.8 & 76.8 & 56.4\\
Uluslar Arası İlişkiler & 58.9 & 55.6 & 49.8 & 35.3 & 80.2 & 55.9\\
Yaşlı Bakımı & 56.3 & 56.8 & 49.0 & 37.9 & 77.7 & 55.5\\
Marka İletişimi & 60.9 & 52.7 & 48.3 & 38.6 & 75.8 & 55.3\\
Laborant ve Veteriner Sağlık & 64.3 & 55.6 & 43.5 & 34.3 & 77.8 & 55.1\\
İnsan Kaynakları Yönetimi & 63.8 & 57.0 & 39.6 & 35.7 & 78.7 & 55.0\\
Çalışma Ekonomisi ve Endüstri İlişkileri & 64.3 & 52.2 & 39.1 & 33.8 & 84.1 & 54.7\\
Halkla İlişkiler ve Reklamcılık & 62.1 & 56.8 & 45.6 & 28.6 & 79.6 & 54.6\\
Emlak ve Emlak Yönetimi & 60.4 & 56.5 & 42.0 & 34.8 & 78.7 & 54.5\\
Felsefe & 59.4 & 54.1 & 44.4 & 35.3 & 79.2 & 54.5\\
Dış Ticaret & 63.1 & 51.9 & 44.7 & 33.5 & 77.7 & 54.2\\
AUZEF & 62.8 & 58.5 & 40.6 & 34.3 & 73.4 & 53.9\\
Spor Yönetimi & 60.9 & 47.8 & 42.5 & 36.7 & 81.6 & 53.9\\
Çocuk Gelişimi & 57.5 & 54.1 & 44.9 & 35.7 & 77.3 & 53.9\\
Lojistik & 58.0 & 50.2 & 43.5 & 36.2 & 80.7 & 53.7\\
Fotoğrafçılık ve Kameramanlık & 58.9 & 54.1 & 42.0 & 30.9 & 76.8 & 52.6\\
Aşçılık & 62.8 & 54.1 & 39.6 & 26.6 & 78.7 & 52.4\\
Halkla İlişkiler ve Tanıtım & 58.5 & 53.6 & 42.5 & 31.9 & 75.4 & 52.4\\
Tarih & 58.5 & 53.6 & 34.3 & 32.9 & 82.6 & 52.4\\
Adalet & 58.7 & 52.4 & 36.4 & 33.5 & 80.6 & 52.3\\
Ev İdaresi & 55.1 & 52.2 & 43.5 & 32.9 & 73.9 & 51.5\\
Muhasebe ve Vergi Uygulamaları & 57.5 & 53.1 & 38.6 & 28.5 & 79.2 & 51.4\\
YGS Denemeleri & 58.5 & 48.3 & 37.7 & 25.6 & 84.5 & 50.9\\
Uluslararası Ticaret ve Lojistik Yönetimi & 55.1 & 51.2 & 43.0 & 28.0 & 70.0 & 49.5\\
DHBT & 57.0 & 52.7 & 31.9 & 20.8 & 80.7 & 48.6\\
Futbol & 55.6 & 48.8 & 30.4 & 22.7 & 84.1 & 48.3\\
Türk Dili ve Edebiyatı & 51.7 & 44.4 & 31.9 & 24.6 & 78.3 & 46.2\\
KPSS Denemeleri & 51.2 & 43.0 & 30.9 & 19.8 & 76.3 & 44.3\\
KPSS & 46.1 & 39.3 & 24.8 & 16.5 & 77.2 & 40.8\\
\midrule
Macro average & 63.6 & 57.5 & 44.4 & 34.8 & 81.4 & 56.4\\
\bottomrule\end{tabular}

\endgroup
\caption{Source-key accuracy (\%) by section, sorted by the descriptive mean over the five runs. Each section has 206 or 207 questions. Labels omit \code{-mlx}; \code{gemma4:26b} is the selector model. Section names follow the source collection. The mean combines different backends and is used only to order rows.}
\label{tab:full-sections}
\end{table*}

\section{Reproducibility and sensitivity checks}\label{app:repro}

For context, a stylized model that knows a fixed fraction of answers and otherwise guesses uniformly would convert the observed five-choice score to 92.89\% with ten choices, a decrease of 0.79 points. The observed decrease is much larger than this reference prediction. However, partial knowledge and nonuniform elimination violate that model, so it does not identify how much loss is caused by guessing, ambiguity, or presentation changes. Likewise, the unpaired full-benchmark scores of 93.35\% and 81.38\% cannot isolate augmentation because their item compositions differ.

The local runs accumulate 1.13--11.56 hours of request time, including local inference and request overhead. The API run records 2.20 million input and 0.33 million output tokens. Using the rates configured in its saved run, the estimated token charge is US\$19.33; this is not an audited invoice. The recorded interval from first to last answer is 0.81 hours, including batch processing and queueing, and is not directly comparable with accumulated local request time.

From the project root, \code{python paper/summarize.py} verifies the pinned source and selected-set reconstruction without model calls. \code{python paper/pilot\_results.py} checks exact response coverage, recomputes scorer-v2 indices and correctness, verifies summary totals and pilot pair counts, reconstructs shared original options, and generates the tables and numerical snapshot. It rejects duplicate response IDs and incompatible full-run conditions. Input hashes include the benchmark, provenance, calibration file, reported-run settings and response files, and analysis code. Paper prose must be reconciled with this snapshot after any new run; table regeneration does not update prose automatically.

Excluding the 100 calibration items gives accuracies of 63.65\%, 57.49\%, 44.43\%, 34.80\%, 81.39\%, in the model order of Table~\ref{tab:full-run}, over 11,900 questions. Each differs from its full-corpus score by less than 0.05 percentage points. This post hoc check removes direct item overlap but not protocol-selection effects.

Among the 981 paired questions, 973 change the relative order of the original options and 892 change the displayed gold label. A post hoc sensitivity analysis resamples whole sections (55 represented sections; 10,000 replicates; NumPy default random generator, seed 42). Its percentile 95\% interval for the paired change is $[-12.89, -8.36]$ percentage points. This permits within-section dependence but assumes exchangeable, independent sections; it is a sensitivity analysis, not a claim of population-representative sampling.

The abbreviated SHA-256 hashes are \code{6cb7c31e...e83ab} for selected data, \code{97dd73cc...cff57} for provenance, and \code{f203ab8e...6686} for calibration data. Full strings and source revisions are recorded in \path{research/artifact_snapshot.json} and \path{research/pilot_snapshot.json}. The latter identifies all five full runs and the five-choice comparison run. \code{paper/check\_audit\_corrections.py} verifies each correction against its recorded original value. \code{paper/audit\_results.py} produces the audit tables and a text-free numeric supplement that reproduces the weighted summaries without exposing question text. Audit inputs and analysis hashes are recorded separately from the unchanged model-response snapshot. The source audit, bibliographic notes, and human-audit rubric are retained under \code{research/}. These files document the analysis; their existence locally does not imply public release of raw responses.

\section{English glosses of source section labels}\label{app:glosses}

Table~\ref{tab:section-glosses} translates the inherited section names for readers of the English manuscript. The original labels remain authoritative identifiers in the frozen dataset and section-level results.

\begin{table*}[t]
\centering
\footnotesize
\setlength{\tabcolsep}{4pt}
\begin{tabular}{@{}ll@{}}\toprule
Source label & English gloss\\\midrule
Ehliyet & Driving licence\\
Dini Bilgiler & Religious knowledge\\
Kim 500 Milyar İster & Who Wants 500 Billion (quiz title)\\
Turizm ve Otel İşletmeciliği & Tourism and hotel management\\
TUS & Medical specialty examination\\
İşletme Yönetimi & Business management\\
Radyo ve Televizyon Programcılığı & Radio and television programming\\
Sosyal Hizmet & Social work\\
Tıbbi Dökümantasyon ve Sekreterlik & Medical documentation and secretarial studies\\
İktisat & Economics\\
Siyer & Biography of the Prophet Muhammad\\
Medya ve İletişim & Media and communication\\
Sağlık Yönetimi & Health management\\
Sosyal Hizmetler & Social services\\
Sosyoloji & Sociology\\
Sağlık Kurumları İşletmeciliği & Healthcare institution management\\
Elektrik Enerjisi Üretim,İletim ve Dağıtımı & Electric power generation, transmission and distribution\\
Menkul Kıymetler ve Sermaye Piyasası & Securities and capital markets\\
Kamu Yönetimi & Public administration\\
Bankacılık ve Sigortacılık & Banking and insurance\\
Yerel Yönetimler & Local government\\
Kültürel Miras ve Turizm & Cultural heritage and tourism\\
İlahiyat & Theology\\
Büro Yönetimi ve Yönetici Asistanlığı & Office management and executive assistance\\
Tarım & Agriculture\\
Turizm ve Seyehat Hizmetleri & Tourism and travel services\\
Özel Koruma ve Güvenlik & Private protection and security\\
Çağrı Merkezi Hizmetleri & Call centre services\\
Parakende Satış ve Mağaza Yöneticiliği & Retail sales and store management\\
Yönetim Bİlişim Sistemleri & Management information systems\\
Uluslar Arası İlişkiler & International relations\\
Yaşlı Bakımı & Elderly care\\
Marka İletişimi & Brand communication\\
Laborant ve Veteriner Sağlık & Laboratory technology and veterinary health\\
İnsan Kaynakları Yönetimi & Human resource management\\
Çalışma Ekonomisi ve Endüstri İlişkileri & Labour economics and industrial relations\\
Halkla İlişkiler ve Reklamcılık & Public relations and advertising\\
Emlak ve Emlak Yönetimi & Real estate and property management\\
Felsefe & Philosophy\\
Dış Ticaret & Foreign trade\\
AUZEF & Open and distance education faculty\\
Spor Yönetimi & Sports management\\
Çocuk Gelişimi & Child development\\
Lojistik & Logistics\\
Fotoğrafçılık ve Kameramanlık & Photography and camera operation\\
Aşçılık & Culinary arts\\
Halkla İlişkiler ve Tanıtım & Public relations and promotion\\
Tarih & History\\
Adalet & Justice\\
Ev İdaresi & Household management\\
Muhasebe ve Vergi Uygulamaları & Accounting and tax practice\\
YGS Denemeleri & Higher education entrance practice tests\\
Uluslararası Ticaret ve Lojistik Yönetimi & International trade and logistics management\\
DHBT & Religious services field examination\\
Futbol & Football\\
Türk Dili ve Edebiyatı & Turkish language and literature\\
KPSS Denemeleri & Public personnel selection practice tests\\
KPSS & Public personnel selection examination\\
\bottomrule\end{tabular}
\caption{English glosses of the 58 source section labels, in the order of Table~\ref{tab:full-sections}. Original spellings are retained for traceability; glosses do not merge sections or certify item provenance. Acronyms and quiz titles denote inherited collection labels, not verified examination ownership.}
\label{tab:section-glosses}
\end{table*}

\end{document}